\documentclass[conference, 10pt]{IEEEtran}
\usepackage{cite}
\usepackage{amsmath,amssymb,amsfonts}
\usepackage{algorithmic}
\usepackage{graphicx}
\usepackage{textcomp}
\usepackage{xcolor}
\def\BibTeX{{\rm B\kern-.05em{\sc i\kern-.025em b}\kern-.08em
    T\kern-.1667em\lower.7ex\hbox{E}\kern-.125emX}}

\IEEEoverridecommandlockouts
\IEEEpubid{\makebox[\columnwidth]{978-1-6654-3166-8/21/\$31.00 $\copyright$2021 IEEE \hfill} \hspace{\columnsep}\makebox[\columnwidth]{ }}

\begin{document}

\title{Delving into Macro Placement \\ with Reinforcement Learning}

\author{\IEEEauthorblockN{Zixuan Jiang}
\IEEEauthorblockA{\textit{The University of Texas at Austin}\\
Austin, TX US \\
zixuan@utexas.edu}
\and
\IEEEauthorblockN{Ebrahim Songhori}
\IEEEauthorblockA{\textit{Google} \\
Mountain View, CA US \\
esonghori@google.com}
\and
\IEEEauthorblockN{Shen Wang}
\IEEEauthorblockA{\textit{Google} \\
Mountain View, CA US \\
wsh4346@gmail.com}
\and
\IEEEauthorblockN{Anna Goldie}
\IEEEauthorblockA{\textit{Google} \\
Mountain View, CA US \\
agoldie@google.com}
\and
\IEEEauthorblockN{Azalia Mirhoseini}
\IEEEauthorblockA{\textit{Google} \\
Mountain View, CA US \\
azalia@google.com}
\and
\IEEEauthorblockN{Joe Jiang}
\IEEEauthorblockA{\textit{Google} \\
Mountain View, CA US \\
wenjiej@google.com}
\and
\IEEEauthorblockN{Young-Joon Lee}
\IEEEauthorblockA{\textit{Google} \\
Mountain View, CA US \\
youngjoonlee@google.com}
\and
\IEEEauthorblockN{David Z. Pan}
\IEEEauthorblockA{\textit{The University of Texas at Austin} \\
Austin, TX US \\
dpan@ece.utexas.edu}
}

\maketitle

\begin{abstract}
In physical design, human designers typically place macros via trial and error, which is a Markov decision process.
Reinforcement learning (RL) methods have demonstrated superhuman performance on the macro placement.
In this paper, we propose an extension to this prior work~\cite{mirhoseini2020chip}.
We first describe the details of the policy and value network architecture.
We replace the force-directed method with DREAMPlace for placing standard cells in the RL environment.
We also compare our improved method with other academic placers on public benchmarks.
\end{abstract}

\section{Introduction}
\label{section:introduction}

Physical design is a critical stage in the chip design process.
It transforms an abstract hypergraph representation (a netlist with devices as nodes and interconnections as hyperedges) into a geometric representation (a layout).
Physical design can be treated as an optimization problem,
where the objective is to maximize the performance of the circuit subject to design constraints.

Due to the complexity of physical design, it is generally decomposed into several sub-stages, including floorplanning, placement, routing, etc.
This split introduces a problem that we lack a synergistic view of these sub-stages.
Taking placement as an example, it is standard practice to minimize wirelength as the main objective, ignoring other metrics.
Although there is work focusing on the co-optimization of these stages (e.g., routing congestion aware placement~\cite{replace} and timing driven placement~\cite{pan200821}),
there is still a huge gap between these sub-stages.
As a result, circuit designers must iterate many times through the end-to-end design flow,
in order to achieve design closure in a production chip design setting.
Specifically, designers conduct physical design in a setting, and try another one according to the previous result.
This procedure will continue until the design closure is achieved.

The design process above is essentially a Markov Decision Process (MDP).
The environment, the design automation software or tools, provides feedback for the agent who makes the decision.
Therefore, the physical design problem becomes an optimization problem in MDP, which is a control problem.
Inspired by the observation above and the success of reinforcement learning (RL), RL methods can be applied to the sub-stages mentioned above.

In~\cite{mirhoseini2020chip}, the authors propose to place macros using an RL algorithm.
The agent places one macro at each time step.
After all macros are placed, the environment places the standard cells and provides feedback to the agent.
The learning algorithm adjusts the agent's policy according to this feedback,
just as human designers learn from feedback at various stages of the physical design process.
By balancing exploration and exploitation, the agent can make superhuman decisions on seen and unseen design benchmarks.

In this paper, we analyze and improve the previous work~\cite{mirhoseini2020chip} to address several issues.
We make contributions in the following three perspectives.
First, we describe the implicit approximation in the policy and value network architecture.
What matters is the correlation with the final evaluation environment instead of the absolute reward during the learning process.
Second, regarding the standard cell placer, we replace the force-directed method with DREAMPlace~\cite{dreamplace},
as the analytical placement is the state-of-the-art method and is widely used in commercial tools.
Third, we make comparisons with academic tools on public benchmarks since the previous work only presents experimental results on the private circuit benchmarks.

\section{Related work}
\label{section:related-work}

\subsection{Analytical placement}
Analytical placement is a nonlinear optimization problem, where the decision variables are the locations of movable nodes.
We minimize the total wirelength such that there is no overlap between nodes.
We can also consider other metrics and constraints in this formulation, such as routing congestion metrics and region constraints.

% In the global placement stage, we minimize the wirelength cost subject to density constraints.
% The problem can be written as
% \begin{equation}
% \begin{split}
%     \min_{x, y} \quad & WL(x, y)\\
%     \text{s.t.} \quad & D(x, y) \leq D_t
% \end{split}
% \end{equation}
% where $x, y$ are the vectors representing the horizontal and vertical locations of nodes, $WL$ and $D$ are the wirelength and density functions, $D_t$ is a given target density.
% This constraint optimization is usually solved by penalty method.
% A series of the following unconstrained problems are solved by gradient-based methods.
% \begin{equation}
%     \min_{x, y} \quad WL(x, y) + \lambda D(x, y)\\
% \end{equation}
% $\lambda$ is gradually increasing to satisfy the original constraint.
ePlace/RePlAce~\cite{eplace-ms, replace} is a family of state-of-the-art placement algorithms.
The density of the placement is modeled as an electrostatic system.
In particular, cells are modeled as electric charges, and density penalty is formulated as the potential energy.
Thus, the movable nodes will spread gradually driven by electric force, which is the gradient w.r.t. the density penalty.
DREAMPlace~\cite{dreamplace} uses a machine learning framework to accelerate the computation of the ePlace/RePlAce algorithm.

\subsection{Standard cell placement and mixed-size placement}
There are two types of movable nodes in the placement problem, standard cells and macros.
Standard cells have similar height, and they are often much smaller than macros.
Moreover, the number of standard cells is much larger than that of macros (1 million vs. 100).

Above all, compared with standard cell placement, it is more difficult to place macros, especially large macros.
The reason is that the large macros make the optimization problem more discrete compared with the standard cell placement.
We solve the global placement in a continuous optimization using gradient based method,
such that a small displacement on a macro may induce a large degradation of the circuit performance during the legalization stage.
Circuit designers usually have to tune the location of large macros by hand in practice.

Researchers also propose specified methods to tackle the mixed-size placement~\cite{eplace-ms}.
Macros are often placed and fixed with heuristic legalization method at first.
Standard cell placement follows as a standard operator afterwards.

\subsection{Macro placement with reinforcement learning}
Mirhoseini et al.~\cite{mirhoseini2020chip} propose to place macros using RL methods.
In their work, the RL agent determines the location of macros one by one.
After all macros are placed and fixed, the macro placement solution is delivered to the environment for evaluation.
In the environment, the standard cells are placed using a force-directed method.
Performance metrics, such as wirelength and congestion, are evaluated and provided as the reward feedback to the RL agent.
The RL learning algorithm updates the agent to improve the reward until good results are achieved.

\section{Method}
\label{section:method}

\subsection{Policy and value network structure}
The original work~\cite{mirhoseini2020chip} applies several explicit approximations so that the environment can provide fast feedback, such as
(1) clustering standard cells,
(2) ignoring pin offset when calculating wirelength,
and (3) discretizing the placement canvas.
We validate that these approximations do not hurt the quality of the feedback significantly, yet saving much time.

Other than that, we find that there are many techniques making a trade-off between time and accuracy implicitly,
such as, graph embedding instead of hypergraph embedding, shallow policy and value neural networks rather than a deep one.
The mask is another typical example.
A mask is generated to do the post-processing on the decision made by the RL agent such that the RL agent will not generate overlaps.
However, intuitively,
a good agent should learn to avoid placing macros in a blank grid instead of an occupied one, which means the mask is unnecessary.
Without the mask, it takes a very long time for the agent to learn to avoid placing macros over each other in our experiments, let alone improving the circuit performance. 
Therefore, we prefer to use the mask since it can accelerate the learning process significantly.

The golden rule is to accelerate the interaction between RL agent and environment since the agent can learn from approximation.
It is not necessary to obtain accurate feedback for the RL agent.

\subsection{DREAMPlace as standard cell placer}
A force-directed method is originally used to conduct standard cell placement in~\cite{mirhoseini2020chip}.
We replace it with DREAMPlace, which yields better placement results.
Also, we believe that DREAMPlace is more similar to the algorithms used in commercial tools.
Since the result of the commercial tools will be treated as the ground truth, the environment will provide more accurate feedback for training agents.
Another advantage of DREAMPlace is its fast execution time, which helps to reduce the learning time.
Please note that the DREAMPlace (or the force-directed method) is only used in the environment.
During the final evaluation, the standard cells are placed by the commercial EDA tools.

\section{Experiments}
\label{section:experiment}

\setlength\tabcolsep{3pt}
\begin{table*}[]
\centering
\caption{Statistics and results on ISPD 2015 benchmarks with movable macros.
$\text{Cong}_H$, $\text{Cong}_V$ represents horizontal and vertical congestion.
Macro locations are from three methods.
Results are reported by a commercial EDA tool after global routing.}
\label{table:ispd-2015-results}
\begin{tabular}{c|cccc|ccc|ccc|ccc}
\hline
\textbf{Benchmark} & \textbf{\# Macros} & \multicolumn{1}{l}{\textbf{\# Std Cells}} & \multicolumn{1}{l}{\textbf{Utility}} & \multicolumn{1}{l|}{\textbf{Max Density}} & \multicolumn{3}{c|}{\textbf{Original Location}} & \multicolumn{3}{c|}{\textbf{DREAMPlace}} & \multicolumn{3}{c}{\textbf{Ours}}     \\
                   &                    & \multicolumn{1}{l}{}                     & \multicolumn{1}{l}{}                 & \multicolumn{1}{l|}{}                     & HPWL                  & $\text{Cong}_H$ & $\text{Cong}_V$    & HPWL               & $\text{Cong}_H$ & $\text{Cong}_V$  & HPWL              & $\text{Cong}_H$ & $\text{Cong}_V$ \\ \hline
des\_perf\_a       & 4                  & 108288                                   & 0.72                                 & 0.72                                      & \textbf{2.19E+06}     & 0.07       & 0.24       & 2.46E+06           & 0.08     & 1.96     & 2.34E+06          & 0.86    & 0.45    \\
edit\_dist\_a      & 6                  & 127413                                   & 0.62                                 & 0.62                                      & 5.42E+06              & 16.63      & 16.18      & 5.01E+06           & 11.79    & 13.16    & \textbf{4.76E+06} & 9.02    & 11.2    \\
fft\_a             & 6                  & 30625                                    & 0.74                                 & 1                                         & 1.29E+06              & 0.60       & 0.12       & 1.22E+06           & 0.64     & 0.12     & \textbf{1.16E+06} & 1.81    & 0.04    \\
fft\_b             & 6                  & 30625                                    & 0.74                                 & 1                                         & 1.46E+06              & 4.67       & 0.37       & \textbf{1.43E+06}  & 3.49     & 0.67     & 1.50E+06          & 5.55    & 0.85    \\
matrix\_mult\_a    & 5                  & 149650                                   & 0.77                                 & 0.92                                      & 5.36E+06              & 1.57       & 1.33       & 5.36E+06           & 0.83     & 2.27     & \textbf{4.43E+06} & 0.29    & 0.4     \\
matrix\_mult\_b    & 7                  & 146435                                   & 0.73                                 & 1                                         & 5.31E+06              & 3.16       & 0.58       & 6.24E+06           & 2.79     & 2.33     & \textbf{5.02E+06} & 0.15    & 0.66    \\
matrix\_mult\_c    & 7                  & 146435                                   & 0.73                                 & 1                                         & \textbf{4.99E+06}     & 0.28       & 0.32       & 5.23E+06           & 1.57     & 0.48     & 5.33E+06          & 4.52    & 0.55    \\
pci\_bridge32\_a   & 4                  & 29517                                    & 0.41                                 & 0.41                                      & 5.94E+05              & 0.32       & 0.11       & 5.06E+05           & 0.11     & 0.62     & \textbf{4.82E+05} & 0       & 0       \\
pci\_bridge32\_b   & 6                  & 28914                                    & 0.51                                 & 0.51                                      & \textbf{7.63E+05}     & 0          & 0          & 9.50E+05           & 0        & 0.06     & 7.85E+05          & 0       & 0       \\ \hline
\textbf{ratio}     &                    & \multicolumn{1}{l}{}                     & \multicolumn{1}{l}{}                 & \multicolumn{1}{l|}{}                     & 1.06                  & -          & -          & 1.09               & -        & -        & 1.00              & -       & -       \\ \hline
\end{tabular}
\end{table*}

To make fair comparisons with academic placers, we demonstrate our results on public benchmarks.

The modern mixed-size (MMS) placement benchmarks~\cite{mms} are widely used for macro placement tasks.
However, they are often used to measure the wirelength optimization due to the missing technology information.
Since our method is not designed only to optimize wirelength, we do not focus on these benchmarks.
% For reference, the wirelength of our method is $1.6\%$ worse than the RePlace on MMS benchmarks.
Instead, we make changes to the ISPD 2015 benchmarks and make an evaluation on them.
We first remove all the blockage and area constraints, and then allow every macro movable in the placement area.
The target density is also updated accordingly.
Since most of the academic placers do not support macro orientation optimization,
we fix the macro orientation.
Table~\ref{table:ispd-2015-results} shows the statistics of the 9 edited benchmarks.
There are overlaps in the original 28nm superblue benchmarks so that it is infeasible for us to allow each single macro movable.
We only pick the designs with macros in 65nm technology.

We evaluate three methods on these benchmarks.
The only difference between these methods is the source of macro location,
which comes from (1) original benchmark, (2) DREAMPlace, and (3) our methods, respectively.
A commercial EDA tool is used to complete the following placement and routing for evaluation.
The half perimeter wirelength (HPWL) and congestion after global routing are listed in Table~\ref{table:ispd-2015-results}.
We obtain a better wirelength and congestion on these benchmarks on average.
On design matrix\_mult\_a, we achieve $17\%$ HPWL reduction with a lower congestion.

\setlength\tabcolsep{2pt}
\begin{table}[]
\centering
\caption{Comparisons on force-driected (FD) method and DREAMPlace (DP).
Four confidential benchmarks are used for evaluation.}
\label{table:internal-benchmarks-results}
\begin{tabular}{c|c|ccc|ccc}
\hline
\textbf{Design} & \textbf{\# Movable} & \multicolumn{3}{c|}{\textbf{RL-FD}} & \multicolumn{3}{c}{\textbf{RL-DP}} \\
                & \textbf{Macros}     & HPWL        & $\text{Cong}_H$ & $\text{Cong}_V$   & HPWL        & $\text{Cong}_H$ & $\text{Cong}_V$   \\ \hline
1               & 926                 & 6.85E+06    & 0.01      & 0.02      & \textbf{6.56E+06}    & 0.01      & 0.02      \\
2               & 873                 & 1.69E+07    & 0         & 0.01      & \textbf{1.64E+07}    & 0         & 0.01      \\
3               & 768                 & 1.36E+07    & 0         & 0         & \textbf{1.30E+07}    & 0         & 0         \\
4               & 1026                & 2.46E+07    & 0         & 0.01      & 2.46E+07    & 0         & 0.01      \\ \hline
\textbf{ratio}  &                     & 1.02        &           &           & 1.00        &           &           \\ \hline
\end{tabular}
\end{table}

The public benchmarks used in this paper are relatively simple,
They do not represent industry circuit designs with advanced technologies, which may have much more macros, more complex routabilities and density requirements.
Therefore, we also present experimental results on four confidential benchmarks with hundreds of movable macros, focusing on the comparison between force-directed method and DREAMPlace,
as shown in Table~\ref{table:internal-benchmarks-results}.
With DREAMPlace as the environment, we achieve smaller wirelength, similar congestion, and $7\%$ less learning time.

\section{Conclusion}
We extend the original work~\cite{mirhoseini2020chip} in three perspectives.
First, we provide more details on the motivation and algorithm design.
Second, DREAMPlace is integrated into the original framework.
Finally, we conduct experiments on public benchmarks and make fair comparisons with academic tools.
We demonstrate that RL can achieve better results and help circuit designers.
\label{section:conclusion}

\bibliographystyle{./IEEEtran}
\bibliography{./IEEEabrv, ./reference}

\end{document}